\documentclass[11pt,a4paper]{article}
\usepackage[hyperref]{acl2019}
\usepackage{times}
\usepackage{latexsym}
\usepackage{url}
\usepackage{hhline}
\usepackage{bm}
\usepackage{bbm} 
\usepackage{graphicx}  %Required
\usepackage{tabularx}
\usepackage{multirow}
\usepackage{booktabs} % For formal tables
\usepackage{mathrsfs}
\usepackage{tabulary}
\usepackage{makecell}
\usepackage{amsmath,amsfonts,amssymb}
\usepackage[ruled,linesnumbered]{algorithm2e}  

\usepackage{algorithmic} 
\usepackage{url}
\usepackage{soul}
\usepackage{balance}

\DeclareMathOperator*{\argmax}{arg\,max}

\aclfinalcopy % Uncomment this line for the final submission
\def\aclpaperid{123} %  Enter the acl Paper ID here

\newcommand{\nop}[1]{}
\newcommand{\fromH}[1]{\textcolor{blue}{From Huan: #1}}
\newcommand{\fromZ}[1]{\textcolor{violet}{#1}}
\newcommand{\add}[1]{\textcolor{red}{#1}}

\title{Reinforced Dynamic Reasoning for Conversational Question Generation}

\author{Boyuan Pan$^{1}$\thanks{~~Work done while visiting the Ohio State University.}, Hao Li$^{1}$, Ziyu Yao$^2$, Deng Cai$^{1,3}$, Huan Sun$^2$\\
	$^1$State Key Lab of CAD$\&$CG, Zhejiang University\\
	$^2$The Ohio State University\\
	$^3$Alibaba-Zhejiang University Joint Institute of Frontier Technologies\\
	{\tt $\{$panby, haolics, dcai$\}$@zju.edu.cn}\\
	{\tt$\{$yao.470, sun.397$\}$@osu.edu} \\
}

\begin{document}
\maketitle

\begin{abstract}
{This paper investigates a new task named \textit{Conversational Question Generation} (CQG) which is to generate a question based on a passage and a conversation history (i.e., previous turns of question-answer pairs). CQG is a crucial task for developing intelligent agents that can drive question-answering style conversations or test user understanding of a given passage. Towards that end, we propose a new approach named Reinforced Dynamic Reasoning (ReDR) network, which is based on the general encoder-decoder framework but incorporates a reasoning procedure in a \textit{dynamic} manner to better understand what has been asked and what to ask next about the passage. To encourage producing meaningful questions, we leverage a popular question answering (QA) model to provide feedback and fine-tune the question generator using a reinforcement learning mechanism. Empirical results {on the recently released CoQA dataset} demonstrate the effectiveness of our method in comparison with various baselines and model variants. Moreover, to show the applicability of our method, we also apply it to create multi-turn question-answering conversations for passages in SQuAD.
}
\end{abstract}
\section{Introduction}
\nop{
\fromH{A better flow might be:}

1. What is question generation, why it is important. 

2. Previous work focus on single-turn question generation. This work focuses on conversational question generation and aims to generate coherent questions in sequence. Why this setting is interesting/important. 

3. Challenges and our approach.

4. Evaluation. 

------
}

In this work, we study a novel task of \textit{conversational question generation} (CQG) which is {given a passage and a conversation history (i.e., previous turns of question-answer pairs), to generate the next question.}%to continue the conversation
\nop{to automatically generate an interconnected question given a passage and past conversation history. } 

\begin{figure}[t]
	\begin{center}
		\includegraphics[width=0.5 \textwidth]{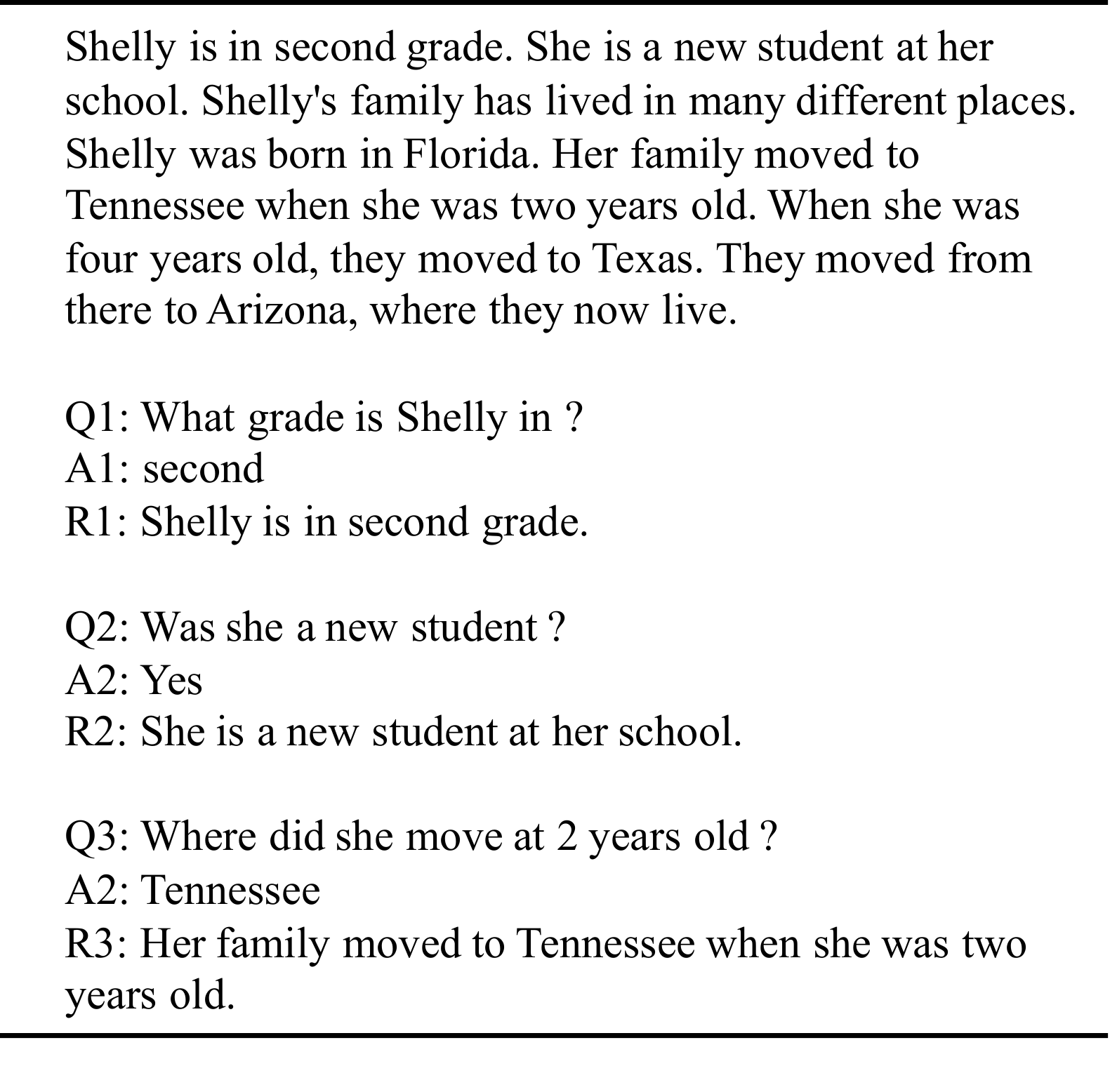}
		\caption{\label{fig1} An example from the CoQA dataset. Each turn contains a question (Q) and an answer (A). The dataset also provides a rationale (R) (i.e., a text span from the passage) to support each answer.}
	\end{center}
\end{figure}

CQG is an important task in its own right for measuring the ability of machines to lead a question-answering style conversation. It can serve as an essential component of intelligent social bots or tutoring systems, asking meaningful and coherent questions to engage users or test student understanding about a certain topic. On the other hand, as shown in Figure \ref{fig1}, large-scale high-quality conversational question answering (CQA) datasets such as CoQA~\cite{reddy2018coqa} and QuAC~\cite{choi2018quac} can help train models to answer sequential questions. However, manually creating such datasets is quite costly, e.g., CoQA spent 3.6 USD per passage on crowdsourcing for conversation collection, and automatic CQG can potentially help reduce the cost, especially when there are a large set of passages available.

In recent years, automatic question generation (QG), which aims to generate natural questions based on a certain type of data sources including structured knowledge bases \cite{serban2016generating,guo2018question} and unstructured texts \cite{rus2010first, heilman2010good, du2017learning, du2018harvesting}, has been widely studied. {However, previous works mainly focus on generating standalone and independent questions based on a given passage. To the best of our knowledge, we are the first to explore CQG, i.e., generating the next question based on a passage and \textit{a conversation history}.}  

Comparing with previous QG tasks, CQG needs to take into account not only the given {passage}, but also the conversation history, and is potentially more challenging as it requires a deep understanding of what has been asked so far and what information should be asked for the next round, in order to make a coherent conversation. 
%while CQG requires to generate a question that is connected with the previous conversation, which potentially brings more challenges

In this paper, we present a novel framework named \emph{Reinforced Dynamic Reasoning} (ReDR) network.\nop{Given a passage and a conversation history of question-answer pairs, our framework first employs a rule-based method to pick up a sentence from the passage as the rationale.} {Inspired by the recent success of reading comprehension models~\citep{xiong2017dynamic,seo2017bidirectional}, ReDR adapts their reasoning procedure (which encodes the knowledge of the passage and the conversation history based on a coattention mechanism) and moreover \textit{dynamically} updates the encoding representation based on a soft decision maker to generate a coherent question.}\nop{Inspired by the recent success of reading comprehension models~\citep{xiong2017dynamic,seo2017bidirectional}, we incorporate the reasoning procedure from them into our Question Generator, which integrates the knowledge of the selected rationale and the conversation history and then dynamically updates the encoding representation to generate a coherent question.} In addition, to encourage ReDR to generate meaningful and interesting questions, ideally, one may employ humans to provide feedback, but as widely acknowledged, involving humans in the loop for training models can be very costly. Therefore, in this paper, we leverage a popular and effective reading comprehension (or QA) model~\cite{chen2017reading} to predict the answer to a generated question and use its answer quality (which can be seen as a proxy for real human feedback) as rewards to fine-tune our model based on a reinforcement learning mechanism~\cite{williams1992simple}. %REINFORCE~\cite{williams1992simple} 

Our contributions are summarized as follows:
%that consists of two components: (1) a rule-based \emph{Rationale Selector} which picks up a sentence of the passage that is most likely to inspire a significant question and (2) a \emph{Dynamic Reasoning Question Generator} that dynamically determines the reasoning depth between the dialog history and the selected rationale. 

\begin{itemize}
	\item We introduce a new task of \textit{Conversational Question Generation} (CQG), which is crucial for developing intelligent agents to drive question-answering style conversations and can potentially provide valuable datasets for future relevant research. 
	%of this task compared to the existing conversational QA and question generation problems.
	
	\item We propose a new and effective framework for CQG, which is equipped with a dynamic reasoning component to generate a conversational question and is further fine-tuned via a reinforcement learning mechanism. %the question generator
	
	\item We show the effectiveness of our method using the recent CoQA dataset. Moreover, we show its wide applicability by using it to create multi-turn QA conversations for passages in SQuAD~\cite{rajpurkar2016squad}.%transferrability and generalizability 
\end{itemize}
\section{Task Definition}

% \nop{\subsection{Conversational Question Answering}
% In the conversational question answering task, we are given a passage $P$ and a converation history contains a series of quesiton-answer pairs $\{(Q_1, A_1), (Q_2, A_2), ..., (Q_{t-1}, A_{t-1})\}$. The goal is to answer the question $Q_t$ in the current turn of the conversation. In CoQA dataset, most answers can be extracted as a subspan of the passages while others are abstractive answers. The CoQA also provides a rationale $R_t$ which is a part of the passage to support each answer, but this is not a necessary component of the task.}

%\subsection{Passage-Aware Conversation Generation}
%\subsection{Task Definition}
% \nop{We define the task of \textit{passage-aware conversational question generation} as: Given a a passage $P$ and a pair of question-answer $(Q_1, A_1)$, and the goal is to extend from it to a multi-turn question-answer conversation $\{(Q_1, A_1), (Q_2, A_2), ..., (Q_{t}, A_{t})\}$. Each turn in the conversation contains a question and an answer.}

%In this section, we define the task of \textit{Conversational Question Generation}. 
Formally, we define the task of \textit{Conversational Question Generation} (CQG) as: Given a passage $X$ and the previous turns of question-answer pairs $\{(q_1, a_1), (q_2, a_2), ..., (q_{k-1}, a_{k-1})\}$ about $X$, CQG aims to generate the next question $q_k$ that is related to the given passage and coherent with the previous questions and answers, i.e.,
\begin{equation}
    q_k = \argmax_{q_k}{P(q_k | X, q_{<k}, a_{<k})}
\end{equation}
where $P(q_k | X, q_{<k}, a_{<k})$ is a conditional probability of generating the question $q_k$.

\nop{Different from previous work on \textit{single-turn} and standalone question generation \cite{du2017learning, zhou2017neural, du2018harvesting} which is solely based on the given sentence or passage, we aim at generating \textit{multi-turn} questions that are coherent extensions to the current conversation, e.g., {possibly contain coreferences to entities mentioned in previous questions and answers.}}
%by bearing coreferences to entities mentioned in the previous questions and answers.

% \subsection{Framework Overview}
% \fromZ{To tackle this task, we propose a question generator based on \textit{Dynamic Reasoning} \cite{?} (Section~\ref{?}). Intuitively, ... put the intuition of dyn reason here...}

% \fromZ{Similar as in conversation generation \cite{sordoni2015neural, serban2016building, li2016diversity, li2016deep}, a typical problem of conversational question generation is that the generator tends to produce trivial and bland questions, such as ``Is he?'' and ``XX'', due to their high occurrence frequencies in the dataset. To address this issue, we further propose a Reinforcement Learning \cite{sutton2018reinforcement} framework to encourage the generator to generate meaningful questions \textit{that are more likely to be answered}. To this end, we adopt a state-of-the-art reading comprehension model \cite{?} to reward the generator when the generated question can be easily answered and punish it otherwise. We elaborate this framework in Section~\ref{?}.}
%\section{Our System: ReDR}
%\section{Framework}
\section{Methodology}
\begin{figure*}[t]
	\begin{center}
		\includegraphics[height=0.4 \textwidth]{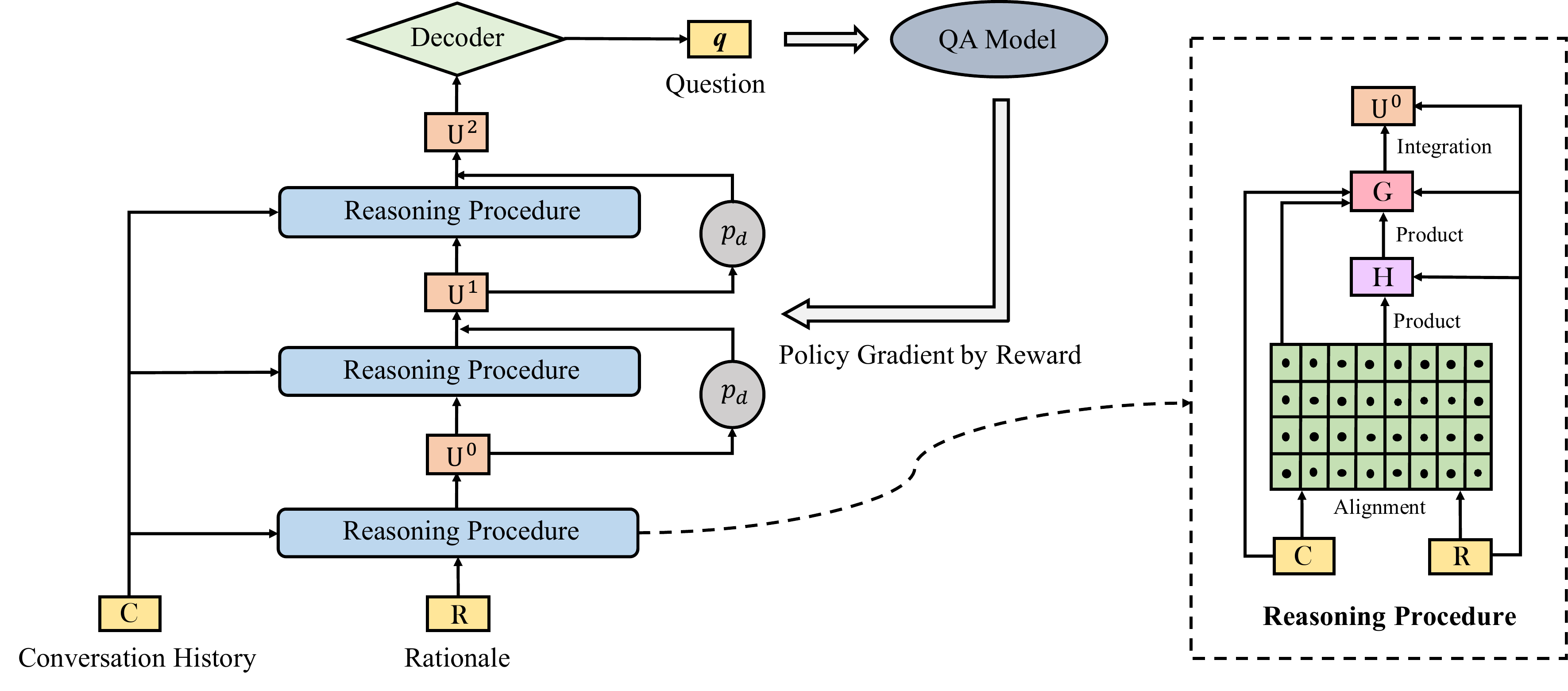}
		\caption{\label{fig2} Overview of our Reinforced Dynamic Reasoning (ReDR) network. The reasoning mechanism iteratively reads the conversation history and at each iteration, its output is dynamically combined with the previous encoding representation through a soft decision maker ($p_d$) as the new encoding representation, which is fed into the next iteration. The model is finally fine-tuned by the reward defined by the quality of the answer predicted from a QA model.\nop{Based on the rationale, the reasoning procedure iteratively reads the conversation history to compute a reasoning vector, which will be combined with the previous encoding representation as the new encoding representation and be fed into the next iteration. }}
	\end{center}
\end{figure*}

We show our proposed framework named \textit{Reinforced Dynamic Reasoning (ReDR)} network in Figure \ref{fig2}. Since a full passage is usually too long and makes it hard to focus on the most relevant information for generating the next question, our method first selects a text span from the passage as the rationale at each conversation turn, %to replace the full passage, Since a full passage is usually too long and hard to inspire a single question, our method first selects a text span of the passage as the rationale at each conversation turn to replace the full passage, 
and then dynamically models the reasoning procedure for encoding the conversation history and the selected rationale, before finally decoding the next question.

\subsection{Rationale Selection}
\label{sec_rs}

We simply set each sentence in the passage as the corresponding rationale for each turn of the conversation.
When experimenting with CoQA, we use the rationale span provided in the dataset. Besides for simplicity and efficiency, another reason that we adopt this rule-based method is that previous research demonstrated that the transition of the dialog attention is smooth~\citep{reddy2018coqa,choi2018quac}, meaning that earlier questions in a conversation are usually answerable by the preceding part of the passage while later questions tend to focus on the ending part of the passage. The selected rationale is then leveraged by subsequent modules for question generation.

% \nop{\add{Most previous work on question generation for reading comprehension \cite{du2017learning, heilman2010good, zhou2017neural} focus on \textit{sentence-level} question generation and assume all sentences in the given passage contain question-worthy concepts. We adopt the same assumption, use this rule-based Rationale Selector for simplicity and efficiency, and leave as an interesting future study designing more complex models like the one in \cite{du2017identifying} for detecting question-worthy sentences in a passage.}}
% \nop{Previous research demonstrated that the transitions of the dialog attention are smooth~\citep{reddy2018coqa,choi2018quac}, which means early questions are mostly answered in the beginning of the passage while later questions tend to focus on the end of the passage. Inspired by this, we first filter out the sentences that contains no name entity (accounted for 38\% of the total number of sentences), and then set the first sentence in the passage as the first rationale and let the rationale at each turn moves to the next sentence. We use this simple and efficient Rationale Selector as our first part of the system, and} The selected rationale is then leveraged by subsequent modules for question generation.

\subsection{Encoding \& Reasoning}
% \nop{In Figure \ref{fig2}, we give an overview of our Question Generator in ReDR. It dynamically models the reasoning procedure for encoding the conversation history and the selected rationale in order to generate the next question.}%\fromH{or rephrase it: It aims to dynamically model the reasoning procedure for generating a question based on the conversation history and the selected rationale.}
% \nop{As presented in Figure \ref{fig2}, we show how our Question Generator dynamically models the reasoning procedure of the conversation history and the selected rationale to generate \nop{the}\fromZ{a} question.\fromZ{From Z: what does the reasoning mean?}}\\~\\
% As mentioned earlier, different from previous question/dialogue generation tasks, we have two knowledge sources (i.e., the conversation history and the rationale) as the inputs. \add{A good encoding of them is crucial for task performance and might involve a reasoning procedure across previous question-answer pairs and the selected rationale for determining the next question}. \nop{The encoding of the inputs is crucial, and a good reasoning procedure \fromH{a few words explaining ``reasoning'', e.g., ``reasoning across multiple question-answer pairs and the selected rationale for determining how the next question connects to them"} may lead to substantial improvements in the final performance. }

At each turn {$k$}, we denote the conversation history\nop{that contains $m$ tokens as} {as a sequence of $m$ tokens}, i.e., $c=\{c_1, c_2, ..., c_m\}$, which concatenates the previous questions and answers $<$$q_1,a_1,...,q_{k-1},a_{k-1}$$>$, and represent the rationale\nop{that contains $n$ tokens as} {as a sequence of $n$ tokens}, i.e., $r=\{r_1, r_2, ..., r_n\}$. As mentioned earlier, different from previous question generation tasks, we have two knowledge sources (i.e., the conversation history and the rationale) as the inputs. A good encoding of them is crucial for task performance and might involve a reasoning procedure across previous question-answer pairs and the selected rationale for determining the next question. We feed them respectively into a bi-directional LSTM and obtain their contextual representations $\mathbf{C}$ $\in$ $R^{d \times m}$ and $\mathbf{R}\in R^{d \times n}$. Inspired by the coattention reasoning mechanism in previous reading comprehension works~\citep{xiong2017dynamic,seo2017bidirectional,pan2017memen}, we compute an alignment matrix of $\mathbf{C}$ and $\mathbf{R}$ to link and fuse the information flow: $\mathbf{S} = \mathbf{R}^{\top} \mathbf{C} \in R^{n \times m}$. We normalize this alignment matrix column-wise (i.e., softmax($\mathbf{S}$)) to obtain the relevance degree of each token in the conversation history to the whole rationale. The new representation of the conversation history w.r.t. the rationale is obtained via:\nop{We normalize this alignment matrix column-wise and multiply it \nop{with}{by} {$\mathbf{R}$} to obtain the relevance degree of each word of the conversation history with the whole rationale:}
\begin{equation}
\begin{aligned}
\label{r1}
\mathbf{H} = \mathbf{R} \cdot {\rm softmax}(\mathbf{S}){\; \in R^{d \times m}}
\end{aligned}
\end{equation}
Similarly, we compute the attention over the conversation history for each word in the rationale via ${\rm softmax}(\mathbf{S}^{\top})$ and obtain the context-dependent representation of the rationale by $\mathbf{C}\cdot{\rm softmax}(\mathbf{S}^{\top})$. In addition, as in \cite{xiong2017dynamic}, we also consider the above new representation of the conversation history and map it to the space of rationale encodings via ${\mathbf{H}\cdot\rm softmax}(\mathbf{S}^{\top})$, and finally obtain the co-dependent representation of the rationale and the conversation history:%map the previous attention vectors into the same encoding space to get the co-dependent representation:
\begin{equation}
\begin{aligned}
\label{r2}
\mathbf{G} = [\mathbf{C};\mathbf{H}] \cdot {\rm softmax}(\mathbf{S}^{\top}) {\; \in R^{2d \times n}}
\end{aligned}
\end{equation}
where $[;]$ means concatenation across row dimension. To deeply capture the interaction between the rationale and the conversation history\nop{among the rationale conditioned on the conversation history\fromZ{between the conversation history and the rationale}}, we feed the co-dependent representation $\mathbf{G}$ combined with the rationale $\mathbf{R}$ into an \textit{integration model} instantiated by\nop{, which is composed of} a bi-directional LSTM:
\begin{equation}
\begin{aligned}
\label{r3}
% \mathbf{u}^0_i = {\rm BiLSTM} ([\mathbf{G}_i;\mathbf{R}_i],\mathbf{u}^0_{i-1})\;{\in R^{\add{3d}}}
\mathbf{u}^0_i = {\rm BiLSTM} (\mathbf{u}^0_{i-1}, \mathbf{u}^0_{i+1}, [\mathbf{G}_i;\mathbf{R}_i]) \in R^d
\end{aligned}
\end{equation}
% \fromZ{unclear:For BiLSTM, would you need both $u_{i-1}^0$ and $u_{i+1}^0$? Is $u_i^0$ a concat of hidden states from both directions? if yes, then why the dimension is 3d, an odd number} 
We define the reasoning process in our paper as Eqn.~(\ref{r1}-\ref{r3}), and now obtain a matrix $\mathbf{U}^0 = [\mathbf{u}^0_1, \mathbf{u}^0_2, ... , \mathbf{u}^0_n]$ as the encoding representation after one-layer reasoning procedure, which can be fed into the decoder subsequently.

\subsection{Dynamic Reasoning}
Oftentimes the conversation history is very informative and complicated, and one single layer of reasoning may be insufficient to comprehend the subtle relationship among the rationale, the conversation history, and the to-be-generated question. Therefore, we propose a dynamic reasoning procedure to iteratively update the encoding representation. We regard $\mathbf{U}^0$ as a new representation of the rationale and input it to the next layer of reasoning together with $\mathbf{C}$:
\begin{equation}
\begin{aligned}
\widetilde{\mathbf{U}}^{1} = F_{reason}(\mathbf{U}^{0}, \mathbf{C})
\end{aligned}
\end{equation}
where $F_{reason}$ is the reasoning procedure (Eqn.~ \ref{r1}-\ref{r3}), and $\widetilde{\mathbf{U}}^{1}$ is the hidden states of the BiLSTM integration model at the next reasoning layer. To effectively learn what information in $\widetilde{\mathbf{U}}^{1}$ and ${\mathbf{U}}^{0}$\nop{ is relevant} is relevant to keep, we use a soft \textit{decision maker} to determine their weights: %of the previous representation and the current representation:
\begin{equation}
\begin{aligned}
%\\
\mathbf{U}^1 &= \mathbf{p}_d \odot \mathbf{U}^0 + (\mathbf{e}_1- \mathbf{p}_d) \odot \widetilde{\mathbf{U}}^{1}\\
\mathbf{p}_d &= \sigma (\mathbf{w}_{\rm u}^{\top}\mathbf{U}^0+ \mathbf{w}_{\rm g}^{\top}\mathbf{G} + \mathbf{w}_{\rm r}^{\top}\mathbf{R} + \mathbf{b})
\end{aligned}
\end{equation}
where $\mathbf{e}_1$ is an all-ones vector, and $\mathbf{w}_u, \mathbf{w}_g, \mathbf{w}_r, \mathbf{b}$ are trainable parameters. {$\mathbf{p}_d \in R^n$ is the decision maker, used as a soft switch to choose between different levels of reasoning.} $\mathbf{U}^{1}$ is the representation to be used for the next layer of reasoning. This iterative procedure halts when a maximum number of reasoning layers $N$ is reached ($N\geq1$)\nop{(\add{which we refer to as \textit{reasoning depth} earlier})}. The final representation $\mathbf{U}^{N}$ is fed into the decoder\nop{, $N$ is the number of the iterative hops}.

\subsection{Decoding}
The decoder generates a word by sampling from the probability $P_{gen}(y_t | y_{<t}, c, r)$ which can be computed via:
\begin{equation}
\begin{aligned}
&P_{gen}(y_t | y_{<t}, c, r) = {\rm MLP}(\mathbf{o}_t, \mathbf{v}_t)\\
&\mathbf{o}_t = {\rm LSTM}(\mathbf{o}_{t-1}, {\rm Emb}(y_{t-1}), \mathbf{v}_{t-1})
\end{aligned}
\end{equation}
\nop{
\begin{equation}
\begin{aligned}
&P_{gen}(y_t | y_{<t}, c, r) = {\rm MLP}(\mathbf{s}_t, \mathbf{c}_t)\\
&\mathbf{s}_t = {\rm LSTM}(\mathbf{s}_{t-1}, {\rm Emb}(y_{t-1}), \mathbf{c}_t)
\end{aligned}
\end{equation}
}
%\fromZ{should we replace $\mathbf{c}_t$ with another symbols? confuse me with $c$ and $c_i$} 
where MLP stands for a standard multilayer perceptron network, $y_t$ is the $t$-th word in the generated question, $\mathbf{o}_t$ is the hidden state of the decoder at time step $t$, and ${\rm Emb}( \cdot )$ indicates the word embedding. $\mathbf{v}_t$ is an attentive read of the encoding representation: $\mathbf{v}_t = \sum_{i=1}^n \alpha_{t,i} \mathbf{u}^N_i$, where the weight $\alpha_{t,i} \in (0,1)$ is scored by another ${\rm MLP}(\mathbf{o}_t,\mathbf{u}^N_i)$ network.

Observing that a question may share common words with the rationale that it is based on and inspired by the widely adopted copy mechanism~\cite{gu2016incorporating,see2017get}, we also apply a pointer network for the generator to copy words from the rationale\nop{input sequence}. Now the probability of generating target word $y_t$ becomes:%in the output sequence 
\begin{equation}
\begin{aligned}
\label{eq_rl}
P(y_t|y_{<t}, c,r) = \lambda P_{gen}(y_t) + ( 1 - \lambda ) P_{pt}(y_t)
\end{aligned}
\end{equation}
where $P_{gen}(y_t)$$=$$P_{gen}(y_t | y_{<t}, c, r)$ is defined earlier, $P_{pt}(y_t) = \sum_{i: r_i = y_t} \alpha_{t,i}$ is the probability of copying word $y_t$ from $r$ (only if $r$ contains $y_t$), and $\lambda$ is the weight to balance the two: %generating the word from the entire vocabulary and copying it from the input rationale:
\begin{equation}
\begin{aligned}
\lambda = \sigma (\mathbf{w}_v^{\top}\mathbf{v}_t + \mathbf{w}_o^{\top}\mathbf{o}_{t} + \mathbf{w}_y^{\top} {\rm Emb}(y_{t-1}) + \mathbf{b}_{pt})
\end{aligned}
\end{equation}
where $\mathbf{w}_v^{\top}$, $\mathbf{w}_o^{\top}$, $\mathbf{w}_y^{\top}$ and $\mathbf{b}_{pt}$ are to be learnt. To optimize all parameters in ReDR, we adopt the maximum likelihood estimation (MLE) approach, i.e., maximizing the summed log likelihood of words in a target question.%In this way, we obtain a question for the new conversation turn.
\nop{
The objective function is to minimize the summed negative log likelihood of the target words. In this way, we obtain a question for the new conversation turn.
}

\subsection{Reinforcement Learning for {Fine-tuning}\nop{Training}}

% \add{remove this paragraph?} Similar as in conversation generation \cite{sordoni2015neural, serban2016building, li2016diversity, li2016deep}, a typical problem of conversational question generation is that the generator tends to produce trivial and bland questions, such as ``what is his name'' and ``what did he do'', due to their high occurrence frequencies in the dataset.\fromZ{Does S2S still face this problem after you correcting the bug?} To address this issue, we further propose a REINFORCE~\cite{williams1992simple} framework as a fine-tuning procedure to encourage the generator to produce meaningful questions \textit{that are more likely to be answered}.\fromH{It seems using RL with this reward cannot directly address the problem. Based on this paragraph, why we use RL is not convincing to me. Suggest the following changes:}

{As shown by recent datasets like CoQA and QuAC, human-created questions tend to be meaningful and interesting. For example, in Figure~\ref{fig1}, given the second rationale R2 ``She is a new student at her school'', humans tend not to ask ``Where is she?'', and similarly given R3, they usually do not create the question ``What happened?''. Although both are legitimate questions, they tend to be less interesting and meaningful compared with the human-created ones shown in Figure~\ref{fig1}. The interestingness or meaningfulness of a question is subjective and hard to define, automatically measuring which is a difficult problem itself. Ideally, one can involve humans in the loop to judge the generated question and provide feedback, but it can be very costly, if not impossible.  }
%and encourage it towards producing meaningful questions

{Driven by such observations, we use the REINFORCE~\cite{williams1992simple} algorithm and adopt one of the state-of-the-art reading comprehension models DrQA~\cite{chen2017reading} as a substitute for humans to provide feedback to the question generator.\nop{is also applied as a strong model for} DrQA answers a question based on the given passage and has {achieved a competitive performance on} CoQA~\cite{reddy2018coqa}. During training, we apply DrQA to answer a generated question, and compare its answer with the human-provided answer (which is associated with the same rationale for generating the question)\footnote{We use the CoQA dataset for training and such information is available as shown in Figure~\ref{fig1}.}. If the answers match well with each other, we regard our generator produces a meaningful question since it asks about the same thing as humans do, and will assign high rewards to such questions.}
%use the score of the answer to reward the generator when the generated question can be easily answered and punish it otherwise. 

Formally, we minimize the negative expected reward for a generated question:
\begin{equation}
\begin{aligned}
J_{RL} = - \mathbb{E}_{q \sim \pi (q | r, c)}[R(a, a^*)]
\end{aligned}
\end{equation}
where $\pi (q | r, c) = \prod_{t} P(y_t|y_{<t}, c,r)$ is the action policy defined in Eqn. (\ref{eq_rl}) for producing question $q$ given rationale $r$ and conversation history $c$, and $R(a, a^*)$ is the reward function defined by the F1 score\footnote{F1 score is the common evaluation metric for QA and is defined as the harmonic mean of precision and recall.} between the DrQA predicted answer $a$ and the human-provided answer $a^*$. For computational efficiency concerns, during training, we make sure that the ground-truth question is in the sampling pool and use beam search to generate 5 more questions. 

{Note that besides providing rewards for fine-tuning our generator, DrQA model also serves another purpose: When applying our framework to any passage,} we can use DrQA to produce an answer to the currently generated question so that the conversation history can be updated for the next-turn of question generation. {In addition, our framework is not limited to DrQA and other more advanced QA models can apply as well.} 
\section{Experiments}
\subsection{Dataset}
We use the CoQA dataset\footnote{https://stanfordnlp.github.io/coqa/}~\cite{reddy2018coqa} to experiment with our ReDR and baseline methods.\nop{Unlike previous reading comprehension datasets, CoQA is a large-scale crowdsourced dataset originally created for building conversational question answering systems.} CoQA contains text passages from diverse domains, conversational questions and answers developed for each passage, as well as rationales (i.e., text spans extracted from given passages) to support answers. The dataset consists of 108k questions in the training set and 8k questions in the development (dev) set with a large hidden test set for competition purpose, and our results are shown on the dev set. \nop{In our setting, we use XXX from their training set for model training and development, and XX from the dev set for testing? When generating a question, we use their associated rationale in the dataset as input, and only incorporate our Rationale Selector into ReDR when applying it to other passages without rationales provided (to be shown later).} %The domain distributions are shown in Table \ref{tab1}.

\begin{table}
	\begin{center}
		\begin{tabular}{lccc}
			\toprule
			\multirow{2}{*}{\textbf{Dataset}}	&  \multirow{2}{*}{\textbf{Passages}} &  \textbf{QA}&  \textbf{Turns per}\\
			 & & \textbf{Pairs} & \textbf{Passage} \\
			\midrule
			Training& 7199 & 10.8k & 15.0 \\ 
			Dev & 500 & 8.0k  & 15.9 \\ 
			\bottomrule
		\end{tabular}
		\vspace{2mm}
	\end{center}
	\caption{\label{tab1} Statistics of the CoQA dataset.}%``Dev" denotes development.       
\end{table}

\begin{table*}[t]
	\centering
	\begin{tabular}{p{6.5cm}cc|cccc}
		\toprule
		\multicolumn{1}{c}{\multirow{2}{*}{\textbf{Models}}} & \multicolumn{2}{c}{\textbf{Relevance}} & \multicolumn{3}{c}{\textbf{Diversity}} &  \\ \cline{2-6} 
		\multicolumn{1}{c}{}  & \textbf{BLEU}          & \textbf{RG-L}    & \textbf{~Dist-1~}     & \textbf{~Dist-2~} &\textbf{~Ent-4~}     \\ \hline
		Vanilla Seq2Seq Model  & 7.64       & 26.68  & 0.010    &  0.034  & 3.370\\
		NQG~\cite{du2017learning}   & 13.97 & 31.75  & 0.017 &  0.068 &  6.518  \\ \hline
		With 1 Layer Reasoning, no RL  & 16.13 & 32.24 & 0.053 & 0.171 & 7.862\\ 
		With 2 Layer Reasoning, no RL & 17.85  & 33.06 & 0.062 & 0.216 & 8.285\\
		With 3 Layer Reasoning, no RL & 17.42  & 32.88 & 0.061 & 0.205 & 8.247 \\
		With Dynamic Reasoning, no RL & 19.10 & 33.57 & 0.064  & 0.220   & 8.304 \\ 
		Reinforced Dynamic Reasoning (ReDR) & \textbf{19.69} & \textbf{34.05}  & \textbf{0.069} & \textbf{0.225}  & \textbf{8.367} \\ 
		\bottomrule
	\end{tabular}
	\vspace{2mm}
	\caption{Quantitative evaluation for conversational question generation using CoQA dataset.}  
	\label{tab2}     
\end{table*}

\subsection{Baselines}
As discussed earlier, CQG has been under-investigated so far, and there are few existing baselines for our comparison. Because of their high relevance with our task as well as their superior performance demonstrated by previous works, we choose to compare with the following models:
\paragraph{Seq2Seq}~\cite{sutskever2014sequence} is a basic encoder-decoder sequence learning system, which has been widely used for machine translation~\cite{luong2015effective} and dialogue generation~\cite{wen2017network}. We concatenate the rationale and the conversation history as the input sequence in our setting. %\add{Seq2seq has been a popular model used for dialogue generation~\cite{} and question generation~\cite{}.}

\paragraph{NQG}~\cite{du2017learning} is a strong attention-based neural network approach for question generation task. The input is the same as the above Seq2Seq model.

\subsection{Implementation Details}
Our word embeddings are initialized by \texttt{glove.840B.300d}~\cite{pennington2014glove}. We set the LSTM hidden unit size to 500 and set the number of layers of LSTMs to 2 in both the encoder and the decoder. Optimization is performed using stochastic gradient descent (SGD), with an initial learning rate of 1.0. The learning rate starts decaying at the step 15000 with a decay rate of 0.95 for every 5000 steps. The mini-batch size for the update is set at 64. We set the dropout~\cite{srivastava2014dropout} ratio as 0.3 and the beam size as 5. The maximum number of iterations for the dynamic reasoning is set to be 3. Since the CoQA contains abstractive answers, we apply DrQA as our question answering model and follow \citet{yatskar2018qualitative} to separately train a binary classifier to produce ``yes" or ``no" for yes/no questions\footnote{Our modified DrQA model achieves 68.8 F1 scores on the CoQA dev set.}. Code is available at \url{https://github.com/ZJULearning/ReDR}.

\nop{\subsection{Baselines}
\paragraph{Seq2Seq}~\cite{sutskever2014sequence} is a basic encoder-decoder sequence learning system for machine translation. We concatenate the rationale and the conversation history as the input sequence.

\paragraph{NQG}~\cite{du2017learning} is a strong attention-based neural networks approach for question generation task. The input is the same as the sequence-to-sequence model. 
\fromZ{add n-layer reasoning and others here?}
}

%\subsection{Results}
\subsection{Automatic Evaluation}

\paragraph{Metrics}~~We follow previous question generation work~\cite{xu2017neural, du2017learning} to use BLEU\footnote{We adopt the 4th smoothing technique as proposed in \cite{chen2014systematic} for short text generation.} \cite{papineni2002bleu} and ROUGE-L \cite{lin2004rouge} to measure the \emph{relevance} between the generated question and the ground-truth one. To evaluate the \emph{diversity} of the generated questions, we follow \cite{li2016diversity} to calculate Dist-n (n=1,2), which is the proportion of unique n-grams over the total number of n-grams in the generated questions for all passages, and \cite{zhang2018generating} to use the Ent-n (n=4) metric, which reflects how evenly the n-gram distribution is {over all generated questions}\nop{\st{in a sentence}\fromH{This is not for each question, right? should be one score for all questions?}}. For all the metrics, the larger they are, the more relevant or diverse {the generated questions are}\nop{the question is}.
%As mentioned by \citet{zhang2018generating}, the Distinct-n metric may lead to biases as it neglects the frequency difference among n-grams. To complement this metric, we additionally

\paragraph{Results and Analysis}~~Table \ref{tab2} shows the performance of various models on the CoQA dataset. As we can see, our model ReDR and its variants perform much better than the baselines, which indicates that the reasoning procedure can significantly boost the quality of the encoding representations and thus improve the question generation performance. 

%In the ablation experiments, 
To investigate the effect of the reasoning procedure and fine-tuning in our model design, we also conduct an ablation study: (1) We first test our model with only one layer of reasoning, i.e., directly feeding the encoding representation $\mathbf{U}^0$ into the decoder. The results drop a lot on all the metrics, which indicates that there is abundant semantic information in the input text so the multi-layer reasoning is necessary. (2) We then augment our model with two or three layers of reasoning but without the decision maker $\mathbf{p}_d$. In other words, we directly use the hidden states of the integration LSTM as the input to the next reasoning layer (formally, $U^j$ = $\tilde{U}^j$). We can see that the performance of our model increases with a two-layer reasoning while decreases with a three-layer reasoning. We conjecture that the two-layer reasoning network is saturated for most of the input text sequences, thus directly adding a layer of network for all the input text seems not optimal. (3) When we add the decision maker to dynamically compute the encoding representations, the results are greatly improved, which demonstrates that using a dynamic procedure can distribute proper weight of each layer to the input sequences in different lengths and amount of information. (4) Finally, we fine-tune the model with the reinforcement learning framework, and the results show that using the answer quality as the reward is helpful for generating better questions.

\begin{table}
	\begin{center}
		\begin{tabular}{lccc}
			\toprule
			 &  \textbf{NQG}&  \textbf{ReDR} & \textbf{Human}\\
			\midrule
			Naturalness  & 1.94  & 1.92 & 2.14\\ 
			Relevance   & 1.16 & 2.02 & 2.82\\ 
			Coherence  & 1.12 & 1.94 & 2.94 \\ 
			Richness  & 1.16 & 2.30 & 2.54 \\ 
			Answerability  & 1.18 & 1.86 & 2.96\\ 
			\bottomrule
		\end{tabular}
		\vspace{2mm}
	\end{center}
	\caption{Human evaluation results on CoQA. ``Human" in the table means the original human-created questions in CoQA.}       
	\label{tab3}
\end{table}

\subsection{Human Evaluation}
We conduct human evaluation to measure the quality of generated questions. We randomly sampled 50 questions along with their conversation history and the passage, and consider 5 aspects: \emph{Naturalness}, which indicates the grammaticality and fluency; \emph{Relevance}, which indicates the connection with the topic of the passage; \emph{Coherence}, which measures whether the generated question is coherent with the conversation history; \emph{Richness}, which measures the amount of information contained in the question. \emph{Answerability}, which indicates whether the question is answerable based on the passage\nop{is easy to answer}. For each sample, 5 people~\footnote{All annotators are native English speakers.} are asked to rank three questions (the ReDR question, the NQG question and the human-created question) by assigning each a score from \{1,2,3\} (the higher, the better). For each aspect, we show the average score across the five annotators on all samples. %ground-truth %people%Ties are allowed?

Table \ref{tab3} shows the results of human evaluation. We can see that our method almost outperforms NQG in all aspects. For \emph{Naturalness}, the three methods obtain the similar scores, which is probably because that the most generated questions are short and fluent, makes them have no significant difference on this aspect. We also observe that on the \emph{Relevance}, \emph{Coherence} and \emph{Answerability} aspects, there is an obvious gap between the generative models and human annotation. This indicates that the contextual understanding is still a challenging problem for the task of the conversational question generation.

\begin{table}
	\begin{center}
		\begin{tabular}{lccc}
			\toprule
			\textbf{Category} &  \textbf{NQG}&  \textbf{ReDR} & \textbf{Human}\\
			\midrule
			\multicolumn{4}{c}{Question Type}\\ \cline{1-4}
			``what" Question  & 0.45  & 0.42 & 0.35\\ 
			``which" Question   & 0.01 & 0.01 & 0.02\\ 
			``when" Question  & 0.07 & 0.05 & 0.04 \\ 
			``where" Question  & 0.08 & 0.06 & 0.07 \\ 
			``who" Question  & 0.06 & 0.22 & 0.15\\ 
			``why" Question   & 0.15 & 0.03 & 0.03 \\ 
			yes/no Question  & 0.08 & 0.07 & 0.21  \\ \hline
			\multicolumn{4}{c}{Linguistic Feature}\\ \cline{1-4}
			Question Length  & 4.05 & 5.34 & 6.48 \\
			Explicit Coref.   & 0.51 & 0.53 & 0.47 \\
			Implicit Coref.  & 0.32 & 0.19 & 0.19 \\ 
			\bottomrule
		\end{tabular}
		\vspace{2mm}
	\end{center}
	\caption{Linguistic statistics for the generated questions and the human annotated questions in CoQA. }       
	\label{tab4}
\end{table}

\subsection{Linguistic Analysis}
We further analyze the generated questions in terms of their linguistic features and constitutions in Table \ref{tab4}, from which we draw three observations: (1) Overall, the distribution of the major types of questions generated by ReDR is closer to human-created questions, in comparison with NQG. For example, ReDR generates a large portion of ``what" and ``who" questions, similarly as humans. (2) We observe that NQG tends to generate many single-word questions such as ``Why?" while our method successfully alleviates this problem. (3) Both ReDR and NQG generate fewer yes/no questions than humans, as a result of generating more ``wh''-type of questions. 

For the relationship between a question and its conversation history, following the analysis in CoQA, we randomly sample 150 questions respectively from each method and observe that about 50\% questions generated by ReDR contain explicit coreference markers such as ``he", ``she" or ``it", which is similar to the other two methods. However, NQG generates much more questions consisting of implicit coreference markers like ``Where?" or ``Who?", which can be less meaningful or not answerable as also verified in Table \ref{tab3}.

\begin{figure}[t]
	\begin{center}
		\includegraphics[width=0.48 \textwidth]{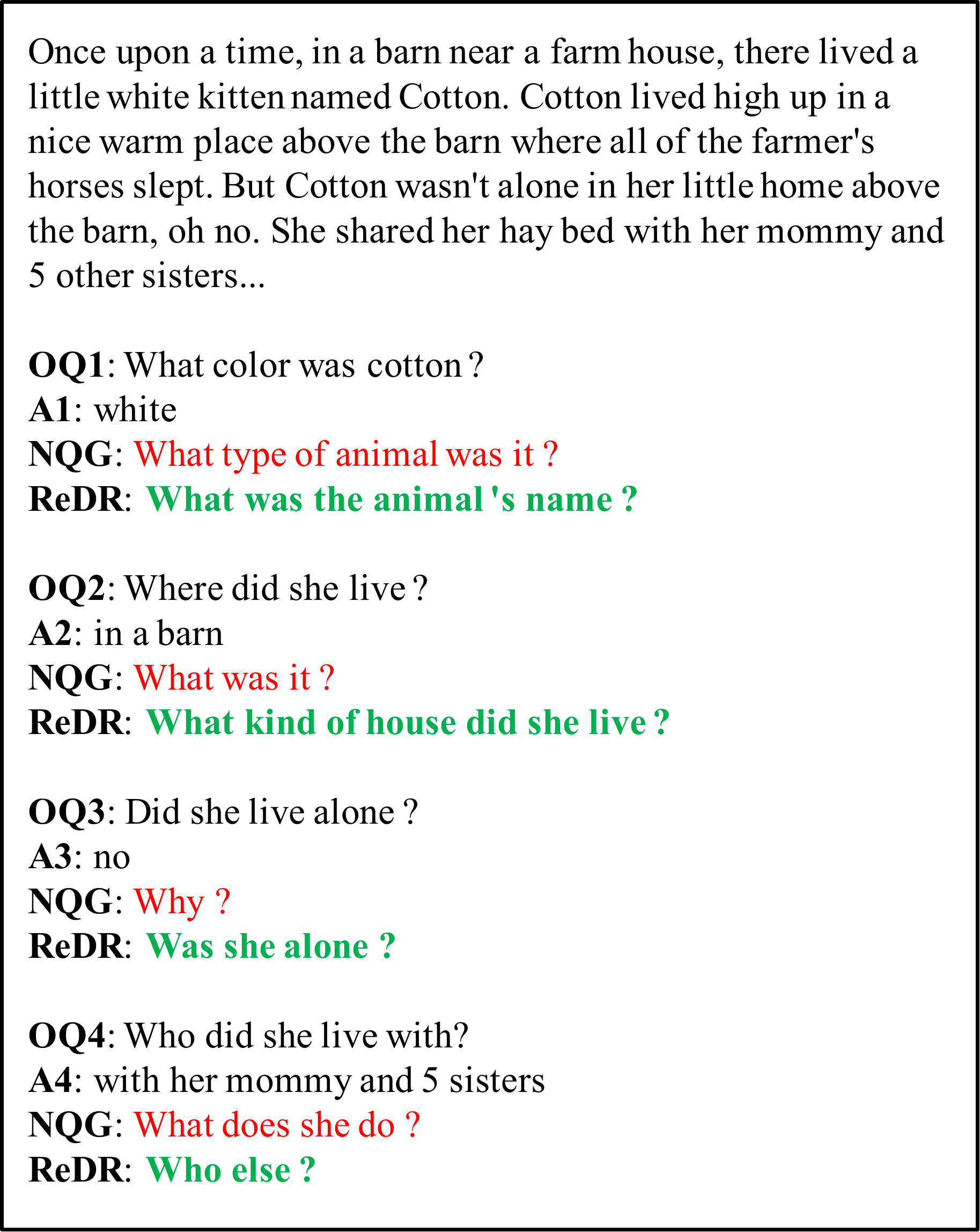}
		\caption{\label{fig3} Example questions generated by human (i.e., original questions denoted as OQ), NQG and our ReDR on CoQA.}
	\end{center}
\end{figure}

\begin{figure}[t]
	\begin{center}
		\includegraphics[width=0.48 \textwidth]{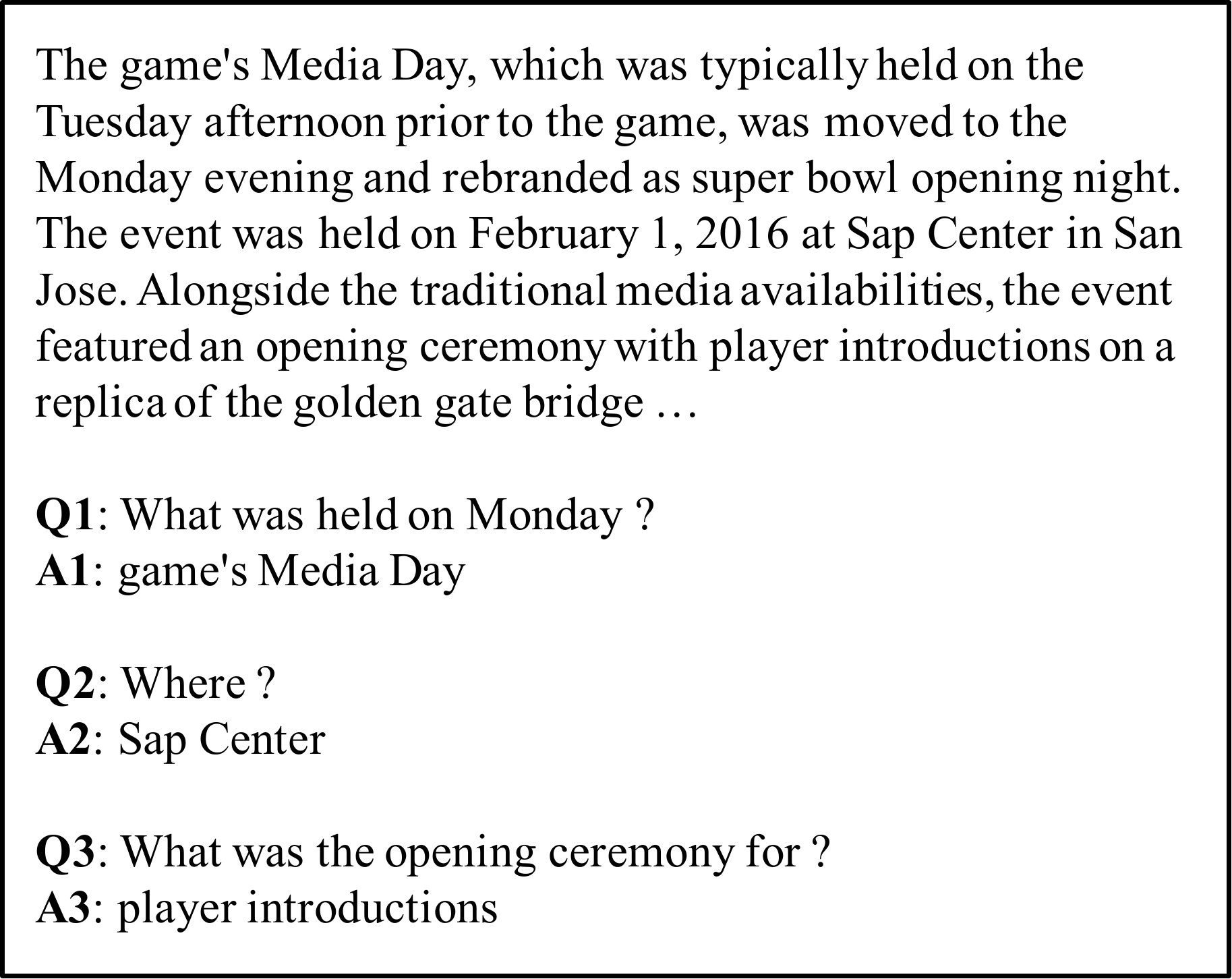}
		\caption{\label{fig4} Our generated conversation on a SQuAD passage. The questions are generated by our ReDR and the answers are predicted by DrQA.}
	\end{center}
\end{figure}

\subsection{Case Study}
%In Figure \ref{fig3}, we show an example sampled from the output questions of our ReDR and NQG on CoQA dataset.
In Figure \ref{fig3}, we show the output questions of our ReDR and NQG on an example from CoQA dataset. For the first turn, both ReDR and NQG generate a meaningful and answerable question. For the second turn, NQG generates ``What was it?", which is answerable and related to the conversation history but simpler than our question ``What kind of house did she live?". For the third turn, NQG generates a coherent but less meaningful question\nop{unanswerable question} ``Why?", while our method generates ``Was she alone?", which is very similar to the human-created question. For the last turn, NQG produces a question that is neither coherent nor answerable, while ReDR asks a much better question ``Who else?".

To show the applicability of ReDR to generate QA style conversations on any passages, we apply it to passages in the SQuAD reading comprehension dataset \cite{rajpurkar2016squad} and show an example in Figure \ref{fig4}. Since there are no rationales provided in the dataset for generating consecutive questions, we first apply our rule-based rationale selection as introduced in Section~\ref{sec_rs} and then generate a question based on the selected rationale and the conversation history. The answers are predicted by our modified DrQA. Figure \ref{fig4} shows that our generated questions are closely related to the passage, e.g., the first question contains ``Monday" and the third one mentions ``opening ceremony". Moreover, we can also generate interesting questions such as ``Where?" which connects to previous questions and makes a coherent conversation.
\section{Related Work}
{\paragraph{Question Generation.} Generating questions from various kinds of sources, such as texts \cite{rus2010first, heilman2010good, mitkov2003computer, du2017learning}, search queries \cite{zhao2011automatically}, knowledge bases \cite{serban2016generating} and images \cite{mostafazadeh2016generating}, has attracted much attention recently. Our work is most related to previous work on generating questions from sentences or paragraphs. Most early approaches are based on rules and templates \cite{heilman2010good, mitkov2003computer}, while \citet{du2017learning} recently proposed to generate a question by a Sequence-to-Sequence neural network model \cite{sutskever2014sequence} with attention \cite{luong2015effective}\nop{where an LSTM encoder \cite{hochreiter1997long} encodes a given sentence into its vector representation and an LSTM decoder generates a question based on the encoding vector}. 
Other approaches such as \cite{zhou2017neural, subramanian2017neural} take into account the answer information in addition to the given sentence or paragraph.}  \cite{du2018harvesting, song2018leveraging} further modeled the surrounding paragraph-level information of the given sentence. However, most of the work focused on generating standalone questions solely based on a sentence or a paragraph. In contrast, this work explores \textit{conversational} question generation and has to additionally consider the conversation history in order to generate a coherent question, making the task much more challenging.

\paragraph{Conversation Generation.} Building chatbots and conversational agents has been pursued by many previous work \cite{ritter2011data, vinyals2015neural, sordoni2015neural, serban2016building, li2016diversity, li2016deep}. \citet{vinyals2015neural} used a Sequence-to-Sequence neural network \cite{sutskever2014sequence} for generating a response given the dialog history. \nop{To avoid generating safe yet {bland}\nop{commonplace} responses like ``I don't know'', \citet{li2016diversity} optimized the generator by maximizing the mutual information between the input sentence and the output generation.}{\citet{li2016diversity} further optimized the response diversity by maximizing the mutual information between inputs and output responses.} Different from these work where the response can be in any form (usually a declarative statement) and is generated solely based on the dialog history, our task is potentially more challenging as it additionally restricts the generated response to be a \textit{follow-up question} about a given \textit{passage}. \nop{To encourage meaningful questions,\nop{unanswerable questions}, we developed a Reinforcement Learning framework and rewarded the generator when the generated question can be easily answered by a reading comprehension model.}

\paragraph{Conversational Question Answering (CQA).} CQA aims to automatically answer a sequence of questions. It has been studied in the knowledge base setting~\cite{saha2018complex,iyyer2017search} and is often framed as a semantic parsing problem. Recently released large-scale datasets~\cite{reddy2018coqa, choi2018quac} enabled studying it in the textual setting where the information source used to answer questions is a given passage, and they inspired many significant work~\cite{zhu2018sdnet,huang2018flowqa,yatskar2018qualitative}. However, collecting such datasets has heavily relied on human efforts and can be very costly. Based on one of the most popular datasets CoQA~\cite{reddy2018coqa}, we examine the possibility of {automatically} \textit{generating} conversational questions, which can potentially reduce the data collection cost for CQA.

% CQA aims to automatically answer a sequence of questions and has been studied under various contexts, e.g., unstructured texts \cite{reddy2018coqa, choi2018quac}, structured knowledge graphs \cite{saha2018complex} and even  semi-structured tables \cite{iyyer2017search}. Several recently released benchmark datasets have inspired many significant works~\cite{zhu2018sdnet,huang2018flowqa,yatskar2018qualitative}. So far collecting such datasets has heavily relied on human efforts and can be very costly. Based on one of the most popular datasets CoQA\cite{reddy2018coqa}, we examine the possibility of {automatically} \textit{generating} conversational questions, which can potentially reduce the data collection cost for CQA.}

\nop{For example, \nop{\citet{iyyer2017search} collected $\sim$6K sequences of simple questions inquiring about tables from Wikipedia. }\citet{choi2018quac} presented $\sim$14K QA dialogs between a student who poses questions about a hidden Wikipedia text and a teacher who extracts relevant text spans from the text to answer the questions. CoQA \cite{reddy2018coqa} is a similar dataset of $\sim$8K question-answering style conversations about open-domain text passages. \fromH{may briefly state the difference between CoQA and QuAC to explain why we use CoQA for model training}}

%and present a new framework for it 
\section{Conclusion}
In this paper, we introduce the task of \emph{Conversational Question Generation} (CQG), and propose a novel framework which achieves promising performance on the popular dataset CoQA. We incorporate a dynamic reasoning procedure to the general encoder-decoder model and dynamically update the encoding representations of the inputs. Moreover, we use the quality of the answers predicted by a QA model as rewards and fine-tune our model via reinforcement learning. In the future, we would like to explore how to better select the rationale for each question. Besides, it would also be interesting to consider using linguistic knowledge such as named entities or part-of-speech tags to improve the coherence of the conversation.
\section{Acknowledgments}
\nop{This work was supported in part by the National Nature Science Foundation of China (Grant Nos: 61751307). Thanks to the anonymous reviewers and Yu Su for their helpful comments and suggestions.}
This research was sponsored in part by the Army Research Office under grant W911NF-17-1-0412, NSF Grant IIS-1815674, the National Nature Science Foundation of China (grant No. 61751307), and Ohio Supercomputer Center \cite{OhioSupercomputerCenter1987}. The views and conclusions contained herein are those of the authors and should not be interpreted as representing the official policies, either expressed or implied, of the Army Research Office or the U.S. Government. The U.S. Government is authorized to reproduce and distribute reprints for Government purposes notwithstanding any copyright notice herein.

\balance
\bibliography{main}

\begin{thebibliography}{44}
\expandafter\ifx\csname natexlab\endcsname\relax\def\natexlab#1{#1}\fi

\bibitem[{Center(1987)}]{OhioSupercomputerCenter1987}
Ohio~Supercomputer Center. 1987.
\newblock \href {http://osc.edu/ark:/19495/f5s1ph73} {Ohio supercomputer
  center}.

\bibitem[{Chen and Cherry(2014)}]{chen2014systematic}
Boxing Chen and Colin Cherry. 2014.
\newblock A systematic comparison of smoothing techniques for sentence-level
  bleu.
\newblock In \emph{Proceedings of the Ninth Workshop on Statistical Machine
  Translation}, pages 362--367.

\bibitem[{Chen et~al.(2017)Chen, Fisch, Weston, and Bordes}]{chen2017reading}
Danqi Chen, Adam Fisch, Jason Weston, and Antoine Bordes. 2017.
\newblock Reading wikipedia to answer open-domain questions.
\newblock In \emph{Proceedings of the 55th Annual Meeting of the Association
  for Computational Linguistics (Volume 1: Long Papers)}, volume~1, pages
  1870--1879.

\bibitem[{Choi et~al.(2018)Choi, He, Iyyer, Yatskar, Yih, Choi, Liang, and
  Zettlemoyer}]{choi2018quac}
Eunsol Choi, He~He, Mohit Iyyer, Mark Yatskar, Wen-tau Yih, Yejin Choi, Percy
  Liang, and Luke Zettlemoyer. 2018.
\newblock Quac: Question answering in context.
\newblock In \emph{Proceedings of the 2018 Conference on Empirical Methods in
  Natural Language Processing}, pages 2174--2184.

\bibitem[{Du and Cardie(2018)}]{du2018harvesting}
Xinya Du and Claire Cardie. 2018.
\newblock Harvesting paragraph-level question-answer pairs from wikipedia.
\newblock In \emph{Proceedings of the 56th Annual Meeting of the Association
  for Computational Linguistics (Volume 1: Long Papers)}, pages 1907--1917.

\bibitem[{Du et~al.(2017)Du, Shao, and Cardie}]{du2017learning}
Xinya Du, Junru Shao, and Claire Cardie. 2017.
\newblock Learning to ask: Neural question generation for reading
  comprehension.
\newblock In \emph{Proceedings of the 55th Annual Meeting of the Association
  for Computational Linguistics (Volume 1: Long Papers)}, volume~1, pages
  1342--1352.

\bibitem[{Gu et~al.(2016)Gu, Lu, Li, and Li}]{gu2016incorporating}
Jiatao Gu, Zhengdong Lu, Hang Li, and Victor~OK Li. 2016.
\newblock Incorporating copying mechanism in sequence-to-sequence learning.
\newblock In \emph{Proceedings of the 54th Annual Meeting of the Association
  for Computational Linguistics (ACL)}, volume~1, pages 1631--1640.

\bibitem[{Guo et~al.(2018)Guo, Sun, Tang, Duan, Yin, Chi, Cao, Chen, and
  Zhou}]{guo2018question}
Daya Guo, Yibo Sun, Duyu Tang, Nan Duan, Jian Yin, Hong Chi, James Cao, Peng
  Chen, and Ming Zhou. 2018.
\newblock Question generation from sql queries improves neural semantic
  parsing.
\newblock In \emph{Proceedings of the 2018 Conference on Empirical Methods in
  Natural Language Processing}, pages 1597--1607.

\bibitem[{Heilman and Smith(2010)}]{heilman2010good}
Michael Heilman and Noah~A Smith. 2010.
\newblock Good question! statistical ranking for question generation.
\newblock In \emph{Human Language Technologies: The 2010 Annual Conference of
  the North American Chapter of the Association for Computational Linguistics},
  pages 609--617. Association for Computational Linguistics.

\bibitem[{Huang et~al.(2018)Huang, Choi, and Yih}]{huang2018flowqa}
Hsin-Yuan Huang, Eunsol Choi, and Wen-tau Yih. 2018.
\newblock Flowqa: Grasping flow in history for conversational machine
  comprehension.
\newblock \emph{arXiv preprint arXiv:1810.06683}.

\bibitem[{Iyyer et~al.(2017)Iyyer, Yih, and Chang}]{iyyer2017search}
Mohit Iyyer, Wen-tau Yih, and Ming-Wei Chang. 2017.
\newblock Search-based neural structured learning for sequential question
  answering.
\newblock In \emph{Proceedings of the 55th Annual Meeting of the Association
  for Computational Linguistics (Volume 1: Long Papers)}, volume~1, pages
  1821--1831.

\bibitem[{Li et~al.(2016{\natexlab{a}})Li, Galley, Brockett, Gao, and
  Dolan}]{li2016diversity}
Jiwei Li, Michel Galley, Chris Brockett, Jianfeng Gao, and Bill Dolan.
  2016{\natexlab{a}}.
\newblock A diversity-promoting objective function for neural conversation
  models.
\newblock In \emph{Proceedings of the 2016 Conference of the North American
  Chapter of the Association for Computational Linguistics: Human Language
  Technologies}, pages 110--119.

\bibitem[{Li et~al.(2016{\natexlab{b}})Li, Monroe, Ritter, Jurafsky, Galley,
  and Gao}]{li2016deep}
Jiwei Li, Will Monroe, Alan Ritter, Dan Jurafsky, Michel Galley, and Jianfeng
  Gao. 2016{\natexlab{b}}.
\newblock Deep reinforcement learning for dialogue generation.
\newblock In \emph{Proceedings of the 2016 Conference on Empirical Methods in
  Natural Language Processing}, pages 1192--1202.

\bibitem[{Lin(2004)}]{lin2004rouge}
Chin-Yew Lin. 2004.
\newblock Rouge: A package for automatic evaluation of summaries.
\newblock \emph{Text Summarization Branches Out}.

\bibitem[{Luong et~al.(2015)Luong, Pham, and Manning}]{luong2015effective}
Thang Luong, Hieu Pham, and Christopher~D Manning. 2015.
\newblock Effective approaches to attention-based neural machine translation.
\newblock In \emph{Proceedings of the 2015 Conference on Empirical Methods in
  Natural Language Processing}, pages 1412--1421.

\bibitem[{Mitkov and Ha(2003)}]{mitkov2003computer}
Ruslan Mitkov and Le~An Ha. 2003.
\newblock Computer-aided generation of multiple-choice tests.
\newblock In \emph{Proceedings of the HLT-NAACL 03 workshop on Building
  educational applications using natural language processing-Volume 2}, pages
  17--22. Association for Computational Linguistics.

\bibitem[{Mostafazadeh et~al.(2016)Mostafazadeh, Misra, Devlin, Mitchell, He,
  and Vanderwende}]{mostafazadeh2016generating}
Nasrin Mostafazadeh, Ishan Misra, Jacob Devlin, Margaret Mitchell, Xiaodong He,
  and Lucy Vanderwende. 2016.
\newblock Generating natural questions about an image.
\newblock In \emph{Proceedings of the 54th Annual Meeting of the Association
  for Computational Linguistics (Volume 1: Long Papers)}, volume~1, pages
  1802--1813.

\bibitem[{Pan et~al.(2017)Pan, Li, Zhao, Cao, Cai, and He}]{pan2017memen}
Boyuan Pan, Hao Li, Zhou Zhao, Bin Cao, Deng Cai, and Xiaofei He. 2017.
\newblock Memen: Multi-layer embedding with memory networks for machine
  comprehension.
\newblock \emph{arXiv preprint arXiv:1707.09098}.

\bibitem[{Papineni et~al.(2002)Papineni, Roukos, Ward, and
  Zhu}]{papineni2002bleu}
Kishore Papineni, Salim Roukos, Todd Ward, and Wei-Jing Zhu. 2002.
\newblock Bleu: a method for automatic evaluation of machine translation.
\newblock In \emph{Proceedings of the 40th annual meeting on association for
  computational linguistics}, pages 311--318. Association for Computational
  Linguistics.

\bibitem[{Pennington et~al.(2014)Pennington, Socher, and
  Manning}]{pennington2014glove}
Jeffrey Pennington, Richard Socher, and Christopher~D. Manning. 2014.
\newblock Glove: Global vectors for word representation.
\newblock In \emph{Empirical Methods in Natural Language Processing (EMNLP)},
  pages 1532--1543.

\bibitem[{Rajpurkar et~al.(2016)Rajpurkar, Zhang, Lopyrev, and
  Liang}]{rajpurkar2016squad}
Pranav Rajpurkar, Jian Zhang, Konstantin Lopyrev, and Percy Liang. 2016.
\newblock Squad: 100,000+ questions for machine comprehension of text.
\newblock In \emph{Proceedings of the 2016 Conference on Empirical Methods in
  Natural Language Processing (EMNLP)}, pages 2383--2392.

\bibitem[{Reddy et~al.(2018)Reddy, Chen, and Manning}]{reddy2018coqa}
Siva Reddy, Danqi Chen, and Christopher~D Manning. 2018.
\newblock Coqa: A conversational question answering challenge.
\newblock \emph{arXiv preprint arXiv:1808.07042}.

\bibitem[{Ritter et~al.(2011)Ritter, Cherry, and Dolan}]{ritter2011data}
Alan Ritter, Colin Cherry, and William~B Dolan. 2011.
\newblock Data-driven response generation in social media.
\newblock In \emph{Proceedings of the conference on empirical methods in
  natural language processing}, pages 583--593. Association for Computational
  Linguistics.

\bibitem[{Rus et~al.(2010)Rus, Wyse, Piwek, Lintean, Stoyanchev, and
  Moldovan}]{rus2010first}
Vasile Rus, Brendan Wyse, Paul Piwek, Mihai Lintean, Svetlana Stoyanchev, and
  Christian Moldovan. 2010.
\newblock The first question generation shared task evaluation challenge.
\newblock In \emph{Proceedings of the 6th International Natural Language
  Generation Conference}.

\bibitem[{Saha et~al.(2018)Saha, Pahuja, Khapra, Sankaranarayanan, and
  Chandar}]{saha2018complex}
Amrita Saha, Vardaan Pahuja, Mitesh~M Khapra, Karthik Sankaranarayanan, and
  Sarath Chandar. 2018.
\newblock Complex sequential question answering: Towards learning to converse
  over linked question answer pairs with a knowledge graph.
\newblock In \emph{Thirty-Second AAAI Conference on Artificial Intelligence}.

\bibitem[{See et~al.(2017)See, Liu, and Manning}]{see2017get}
Abigail See, Peter~J Liu, and Christopher~D Manning. 2017.
\newblock Get to the point: Summarization with pointer-generator networks.
\newblock In \emph{Proceedings of the 55th Annual Meeting of the Association
  for Computational Linguistics (Volume 1: Long Papers)}.

\bibitem[{Seo et~al.(2017)Seo, Kembhavi, Farhadi, and
  Hajishirzi}]{seo2017bidirectional}
Minjoon Seo, Aniruddha Kembhavi, Ali Farhadi, and Hannaneh Hajishirzi. 2017.
\newblock Bidirectional attention flow for machine comprehension.
\newblock \emph{ICLR}.

\bibitem[{Serban et~al.(2016{\natexlab{a}})Serban, Sordoni, Bengio, Courville,
  and Pineau}]{serban2016building}
Iulian~V Serban, Alessandro Sordoni, Yoshua Bengio, Aaron Courville, and Joelle
  Pineau. 2016{\natexlab{a}}.
\newblock Building end-to-end dialogue systems using generative hierarchical
  neural network models.
\newblock In \emph{Thirtieth AAAI Conference on Artificial Intelligence}.

\bibitem[{Serban et~al.(2016{\natexlab{b}})Serban, Garc{\'\i}a-Dur{\'a}n,
  Gulcehre, Ahn, Chandar, Courville, and Bengio}]{serban2016generating}
Iulian~Vlad Serban, Alberto Garc{\'\i}a-Dur{\'a}n, Caglar Gulcehre, Sungjin
  Ahn, Sarath Chandar, Aaron Courville, and Yoshua Bengio. 2016{\natexlab{b}}.
\newblock Generating factoid questions with recurrent neural networks: The 30m
  factoid question-answer corpus.
\newblock In \emph{Proceedings of the 54th Annual Meeting of the Association
  for Computational Linguistics (Volume 1: Long Papers)}, volume~1, pages
  588--598.

\bibitem[{Song et~al.(2018)Song, Wang, Hamza, Zhang, and
  Gildea}]{song2018leveraging}
Linfeng Song, Zhiguo Wang, Wael Hamza, Yue Zhang, and Daniel Gildea. 2018.
\newblock Leveraging context information for natural question generation.
\newblock In \emph{Proceedings of the 2018 Conference of the North American
  Chapter of the Association for Computational Linguistics: Human Language
  Technologies, Volume 2 (Short Papers)}, volume~2, pages 569--574.

\bibitem[{Sordoni et~al.(2015)Sordoni, Galley, Auli, Brockett, Ji, Mitchell,
  Nie, Gao, and Dolan}]{sordoni2015neural}
Alessandro Sordoni, Michel Galley, Michael Auli, Chris Brockett, Yangfeng Ji,
  Margaret Mitchell, Jian-Yun Nie, Jianfeng Gao, and Bill Dolan. 2015.
\newblock A neural network approach to context-sensitive generation of
  conversational responses.
\newblock In \emph{Proceedings of the 2015 Conference of the North American
  Chapter of the Association for Computational Linguistics: Human Language
  Technologies}, pages 196--205.

\bibitem[{Srivastava et~al.(2014)Srivastava, Hinton, Krizhevsky, Sutskever, and
  Salakhutdinov}]{srivastava2014dropout}
Nitish Srivastava, Geoffrey~E Hinton, Alex Krizhevsky, Ilya Sutskever, and
  Ruslan Salakhutdinov. 2014.
\newblock Dropout: a simple way to prevent neural networks from overfitting.
\newblock \emph{Journal of Machine Learning Research}, 15(1):1929--1958.

\bibitem[{Subramanian et~al.(2017)Subramanian, Wang, Yuan, Zhang, Bengio, and
  Trischler}]{subramanian2017neural}
Sandeep Subramanian, Tong Wang, Xingdi Yuan, Saizheng Zhang, Yoshua Bengio, and
  Adam Trischler. 2017.
\newblock Neural models for key phrase detection and question generation.
\newblock \emph{arXiv preprint arXiv:1706.04560}.

\bibitem[{Sutskever et~al.(2014)Sutskever, Vinyals, and
  Le}]{sutskever2014sequence}
Ilya Sutskever, Oriol Vinyals, and Quoc~V Le. 2014.
\newblock Sequence to sequence learning with neural networks.
\newblock In \emph{Advances in neural information processing systems}, pages
  3104--3112.

\bibitem[{Vinyals and Le(2015)}]{vinyals2015neural}
Oriol Vinyals and Quoc Le. 2015.
\newblock A neural conversational model.
\newblock \emph{arXiv preprint arXiv:1506.05869}.

\bibitem[{Wen et~al.(2017)Wen, Vandyke, Mrk{\v{s}}i{\'c}, Gasic, Barahona, Su,
  Ultes, and Young}]{wen2017network}
Tsung-Hsien Wen, David Vandyke, Nikola Mrk{\v{s}}i{\'c}, Milica Gasic, Lina
  M~Rojas Barahona, Pei-Hao Su, Stefan Ultes, and Steve Young. 2017.
\newblock A network-based end-to-end trainable task-oriented dialogue system.
\newblock In \emph{Proceedings of the 15th Conference of the European Chapter
  of the Association for Computational Linguistics: Volume 1, Long Papers},
  volume~1, pages 438--449.

\bibitem[{Williams(1992)}]{williams1992simple}
Ronald~J Williams. 1992.
\newblock Simple statistical gradient-following algorithms for connectionist
  reinforcement learning.
\newblock \emph{Machine Learning}, 8(3-4):229--256.

\bibitem[{Xiong et~al.(2017)Xiong, Zhong, and Socher}]{xiong2017dynamic}
Caiming Xiong, Victor Zhong, and Richard Socher. 2017.
\newblock Dynamic coattention networks for question answering.
\newblock \emph{ICLR}.

\bibitem[{Xu et~al.(2017)Xu, Liu, Wang, Chengjie, Wang, Wang, and
  Qi}]{xu2017neural}
Zhen Xu, Bingquan Liu, Baoxun Wang, SUN Chengjie, Xiaolong Wang, Zhuoran Wang,
  and Chao Qi. 2017.
\newblock Neural response generation via gan with an approximate embedding
  layer.
\newblock In \emph{Proceedings of the 2017 Conference on Empirical Methods in
  Natural Language Processing}, pages 617--626.

\bibitem[{Yatskar(2018)}]{yatskar2018qualitative}
Mark Yatskar. 2018.
\newblock A qualitative comparison of coqa, squad 2.0 and quac.
\newblock \emph{arXiv preprint arXiv:1809.10735}.

\bibitem[{Zhang et~al.(2018)Zhang, Galley, Gao, Gan, Li, Brockett, and
  Dolan}]{zhang2018generating}
Yizhe Zhang, Michel Galley, Jianfeng Gao, Zhe Gan, Xiujun Li, Chris Brockett,
  and Bill Dolan. 2018.
\newblock Generating informative and diverse conversational responses via
  adversarial information maximization.
\newblock In \emph{Advances in Neural Information Processing Systems}, pages
  1815--1825.

\bibitem[{Zhao et~al.(2011)Zhao, Wang, Li, Liu, and
  Guan}]{zhao2011automatically}
Shiqi Zhao, Haifeng Wang, Chao Li, Ting Liu, and Yi~Guan. 2011.
\newblock Automatically generating questions from queries for community-based
  question answering.
\newblock In \emph{Proceedings of 5th international joint conference on natural
  language processing}, pages 929--937.

\bibitem[{Zhou et~al.(2017)Zhou, Yang, Wei, Tan, Bao, and
  Zhou}]{zhou2017neural}
Qingyu Zhou, Nan Yang, Furu Wei, Chuanqi Tan, Hangbo Bao, and Ming Zhou. 2017.
\newblock Neural question generation from text: A preliminary study.
\newblock In \emph{National CCF Conference on Natural Language Processing and
  Chinese Computing}, pages 662--671. Springer.

\bibitem[{Zhu et~al.(2018)Zhu, Zeng, and Huang}]{zhu2018sdnet}
Chenguang Zhu, Michael Zeng, and Xuedong Huang. 2018.
\newblock Sdnet: Contextualized attention-based deep network for conversational
  question answering.
\newblock \emph{arXiv preprint arXiv:1812.03593}.

\end{thebibliography}
\bibliographystyle{acl_natbib}

\end{document}